\documentclass{article}

\usepackage[nonatbib, final]{neurips_2023}
\usepackage[verbose=true,letterpaper]{geometry}
\AtBeginDocument{
  \newgeometry{
    textheight=9in,
    textwidth=6.5in,
    top=1in,
    left=1in,    % Added to ensure 1 inch margin on the left
    right=1in,   % Added to ensure 1 inch margin on the right
    bottom=1in,
    headheight=12pt,
    headsep=25pt,
    footskip=30pt
  }

}

\usepackage[utf8]{inputenc} % allow utf-8 input
\usepackage[T1]{fontenc}    % use 8-bit T1 fonts
\usepackage{hyperref}       % hyperlinks
\usepackage{url}            % simple URL typesetting
\usepackage{multirow} % added
\usepackage{booktabs}       % professional-quality tables
\usepackage{amsfonts}       % blackboard math symbols
\usepackage{nicefrac}       % compact symbols for 1/2, etc.
\usepackage{microtype}      % microtypography
\usepackage{xcolor}         % colors
\usepackage{wrapfig} % added
\usepackage{graphicx}
\usepackage{enumitem} % Add this line
\usepackage{subcaption} % added
\usepackage{amsmath}
\usepackage{mathptmx}
\usepackage{siunitx} % Add this line to use the S column type
\usepackage[numbers]{natbib}

\title{\Large Fairness in Autonomous Driving: Towards Understanding\\ Confounding Factors in Object Detection under Challenging Weather}

\author{%
  Bimsara Pathiraja, Caleb Liu, Ransalu Senanayake\\
  Arizona State University \\
  \texttt{\{bpathir1,calebliu,ransalu\}}@asu.edu
}

\begin{document}

\maketitle

\normalsize

\begin{abstract}

The deployment of autonomous vehicles (AVs) is rapidly expanding to numerous cities. At the heart of AVs, the object detection module assumes a paramount role, directly influencing all downstream decision-making tasks by considering the presence of nearby pedestrians, vehicles, and more. Despite high accuracy of pedestrians detected on held-out datasets, the potential presence of algorithmic bias in such object detectors, particularly in challenging weather conditions, remains unclear. This study provides a comprehensive empirical analysis of fairness in detecting pedestrians in a state-of-the-art transformer-based object detector. In addition to classical metrics, we introduce novel probability-based metrics to measure various intricate properties of object detection. Leveraging the state-of-the-art FACET dataset and the Carla high-fidelity vehicle simulator, our analysis explores the effect of protected attributes such as gender, skin tone, and body size on object detection performance in varying environmental conditions such as ambient darkness and fog. Our quantitative analysis reveals how the previously overlooked yet intuitive factors, such as the distribution of demographic groups in the scene, the severity of weather, the pedestrians' proximity to the AV, among others, affect object detection performance. Our code is available at \url{https://github.com/bimsarapathiraja/fair-AV}.

\end{abstract}

\section{Introduction}

Recent advances in large-scale machine learning (ML) models have significantly contributed to the development of semi-autonomous systems. For instance, we already see object detection algorithms integrated into driver assistant systems of most modern automobiles. However, as these systems evolve to make fully autonomous decisions around humans, it becomes crucial to understand their effectiveness in different scenarios. In an era where autonomous vehicles (AVs) are proliferating in numerous cities, the stakes of failure are high. To study these effects, we center our discussion around the effect of ML on humans in the context of AVs. 

When an AV operates in an urban environment, the very first step is identifying objects around it using an object detector. These objects are then used by other downstream modules such as trajectory prediction, path planning, and control. Hence, a minor error in object detection can magnify within these downstream modules, leading to collisions. For instance, in a fatal accident that happened in Tempe, Arizona, the object detector failed to correctly identify a person, making the downstream planning algorithm ignore the person~\cite{Madrigal.2018}.

Object detectors exhibit two primary failure modes: the failure to detect objects and the hallucination of non-existent objects. However, there is a lack of understanding on 1) whether these failures tend to occur predominantly for specific demographic groups and 2) the impact of these failures on those groups under varying operational conditions. For example, when RGB cameras are used for object detection, it can be hypothesized that individuals with different skin tones and body sizes may be affected disparately by detection failures. It can further be hypothesized that specific operational conditions, such as ambient darkness or fog, may have differential effects on these demographic groups.

Despite the myriad ways in which an object detector in AVs can potentially harm different demographic groups, there is very limited research on characterizing algorithmic bias in such systems. Consequently, there are only few metrics available to assess failure modes in object detection, because most fairness metrics are developed primarily for classification tasks. State-of-the-art (SOTA) object detectors share similarities with image classification but also exhibit major differences. One commonality is that object detectors can be conceptualized as region-level classifiers with some additional functionalities. In fact, many object detection algorithms utilize image classification neural network architectures as their backbones~\cite{girshick2015fast, carion2020endtoend}. Apart from generating region-level object class predictions, object detectors also produce many potential bounding boxes that represent the boundaries of the detected objects. These bounding boxes are filtered in a subsequent step. As we will discuss in Section~\ref{sec:fair_def}, due to these differences, certain existing fairness metrics developed for classification do not fully represent intricacies of object detection in AVs.

In this work, we propose fairness metrics tailored for evaluating object detection in AVs, with a specific focus on enhancing \emph{equal opportunity}. While engineers should diligently work to minimize potential failures, reducing the failure rate to zero is impractical for any deep learning model~\cite{Sagar2024icml}. Acknowledging the inevitability of failures in object detection, our goal is to assess whether two demographic groups are treated equally by the object detection neural network under various weather conditions. We compute metrics based on the correct and incorrect predictions of bounding boxes. Some of the metrics are based on the number of detections, making them similar to conventional fairness metrics. Additionally, we also define new metrics based on the probabilities associated with predicted bounding boxes as uncertainty~\cite{Senanayake2024arxiv_unc} plays a major role in safety-critical systems such as AVs.

We use the metrics to evaluate one of the best-performing object detectors, Detection Transformers (DETR)~\cite{carion2020endtoend}. We also use the recent FACET dataset~\cite{gustafson2023facet} as it contains explicit skin tone annotations. Since it is impossible to replicate the same environment with the same pedestrians under varying weather conditions, we resort to Carla~\cite{dosovitskiy2017carla}, a high-fidelity simulator, in whcih we can control the weather and pedestrians to quantify how weather affects different \emph{protected groups}. Additionally, such simulators allow us to change different protected attributes of a person including skin tone, body size, and gender will allow counterfactual analysis.

We conducted experiments in foggy and darker weather conditions for pedestrians with varying skin tones, genders, and body sizes. Our results show that the skin tone of a person and the body size, depending on the distribution of people who belong to these demographic groups in the scene, disparately affect detection performance. Results also indicate how different demographic groups are affected (or not affected) by the severity of weather conditions and the distance between the vehicle and pedestrians.

Simulation helped us reveal fairness disparities which otherwise would not have been able to understand using a limited dataset. These results highlight the importance of testing the computer vision pipeline of AVs in both real-world datasets and controlled-simulated environments. We advocate the development of such test beds to audit fairness in AVs and other autonomous decision-making systems that rely on images as they help engineers with debugging models and legal authorities to set legal guidelines. We summarize our key contributions as follows:
\begin{enumerate}
  \item We propose confidence score-based metrics for fairness evaluation to bridge the gap between object detection and fairness evaluation. To our knowledge, this marks the first attempt to formulate such metrics, specifically tailored for fairness assessment in object detection.
  % \item We demonstrate, using results from the real-world dataset, how minor data alterations could invert perceived biases. Such findings underscore the necessity of simulations to unravel the nuanced factors that affects of fairness in object detection.
  \item We conduct comprehensive experiments and present the trends in AV biases for protected attributes under numerous weather conditions utilizing a SOTA transformer-based object detection model.
\end{enumerate}

\section{Background and Related Works}

\subsection{Computer Vision in Autonomous Vehicles}
An AV comprises two major cascaded modules: perception and decision-making. Since the decision-making module relies completely on the signals it obtain from the perception module, maintaining a good fairness and accuracy in perception is crucial to develop safe and ethical AVs~\cite{Kruse2022arxiv}. The perception system is a collection of deep neural networks used for object detection, object tracking, and trajectory prediction~\cite{Delecki2022arxiv}. Since object detection is the first step, any mistake it makes passes down to other downstream modules. While some AVs use both LIDAR and RGB camera, some only use RGB cameras. Irrespective of whether an AV uses a LIDAR or not, RGB cameras play a critical role in AVs as their rich and dense representation of color values significantly improves the performance of computer vision tasks~\cite{paigwar2021frustum}.

\subsection{Fairness in Object Detection}
While the majority of fairness studies focus on tabular data, there are a few studies on fairness in computer vision~\cite{kim2023humans, meister2023gender, fabbrizzi2022survey, hall2023vision, yang2022enhancing}. Some studies have shown that commercial gender classifiers perform better for lighter skin people and for male~\cite{buolamwini2018gender}. There are also a number of datasets for evaluating fairness in image classification~\cite{liu2015deep,daneshjou2022disparities}. Despite the use of object detection in safety critical applications such as autonomous driving, there are only a very few studies on the bias and fairness in such systems. 

Brandao~\cite{brandao2019age} analyzes age and gender bias in pedestrian detection using classical detectors such as support vector machines and AdaBoost. The study reveals a higher object detection miss rate for children and a slightly higher miss rate for females. There have also been attempts to mitigate this performance disparity with age through various data collection techniques~\cite{cmsf2022003011} for variants of classical object detectors~\cite{M2Det2019aaai}. Li et al.~\cite{li2023darkskin} argue that, for older object detectors such as YOLO and Fast-RCNN, pedestrians with darker skin tones and children are more susceptible to misdetection, while gender has no effect, especially at nighttime. Testing on then-cutting-edge models such as Mask-RCNN and Faster-RCNN, ~\cite{wilson2019predictive} also found that the rate of correct predictions is higher for light-skin individuals. 

Our study differs from these other studies in multiple ways. We focus on SOTA transformer-based models such as DETR~\cite{carion2020endtoend} because, considering their perceived superior performance, they are likely to be used by all AVs. Since their attention mechanism~\cite{vaswani2017attention} allows a global view of the image, it is possible that it focuses on the person as a whole rather than a few pixels on the exposed part of the person's skin. As another difference, we do not come to conclusions completely relying on a small annotated dataset collected from the real world as it is impossible to disentangle confounding factors in such images. We also evaluate the effect of weather, such as fog, on different demographic groups, focusing on the distance of pedestrians to the car as a contributing factor for accuracy. To evaluate these effects, instead of relying solely on standard metrics developed for classification, we introduce some additional metrics to perform an in-depth analysis. 

\subsection{The Effect of Weather on Object Detection}
Studies tackle adverse weather in one of two ways: dataset creation or domain adaption~\cite{liu2020large}. In particular, Thopalli et al. ~\cite{Thopalli_2023_ICCV} use annotations to infer context for the image and its instances to improve domain adaption. Kenk et al.~\cite{kenk2020dawn} create a dataset in diverse traffic and weather environments, including sandstorms, a phenomenon that less commonly appears in weather datasets. In addition, Ding et al.~\cite{ding2023cf} also create a snow-only dataset, as well as a metric to define how much snow obscures the image. 

Notably, many of these studies rely on the mean average precision metric for benchmarking. The use of the probability-based metrics defined in this study could shed more light on the usefulness and feasibility of these methods. Previous studies does not explicitly study the effect of many weather conditions on different demographic groups. Our study also differs from others by using a controlled simulator to produce synthetic data rather than artificially removing rain~\cite{rs15061487} or fog~\cite{Shen_Yu_Shu_Qin_Wei_2023} for object detection. Such simulators can generate physics-based, natural-looking weather conditions~\cite{gawron2018heterogeneous} and allow engineers to set up various experiments.
%  ##################

\section{Metrics}

\subsection{Measuring Correct and Incorrect Predictions in Object Detection}

\begin{figure}
  \begin{minipage}[b]{0.35\textwidth}
    \centering
    \includegraphics[width=0.9\linewidth]{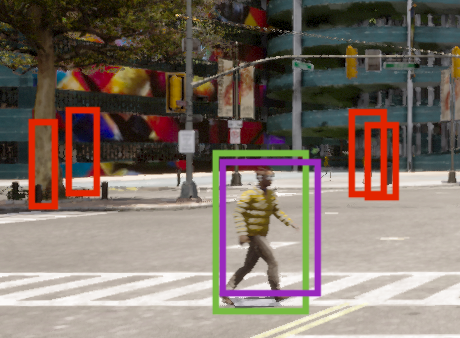}
    \caption{Ground truth for a pedestrian (green), true positive for the pedestrian (purple) and the false positives (red) are shown. The IoU between green and purple bounding boxes is high and the IoUs between green and red bounding boxes are zero.}
    \label{fig:preds}
  \end{minipage}
  \begin{minipage}[b]{0.6\textwidth}
    \centering
    \includegraphics[width=\linewidth]{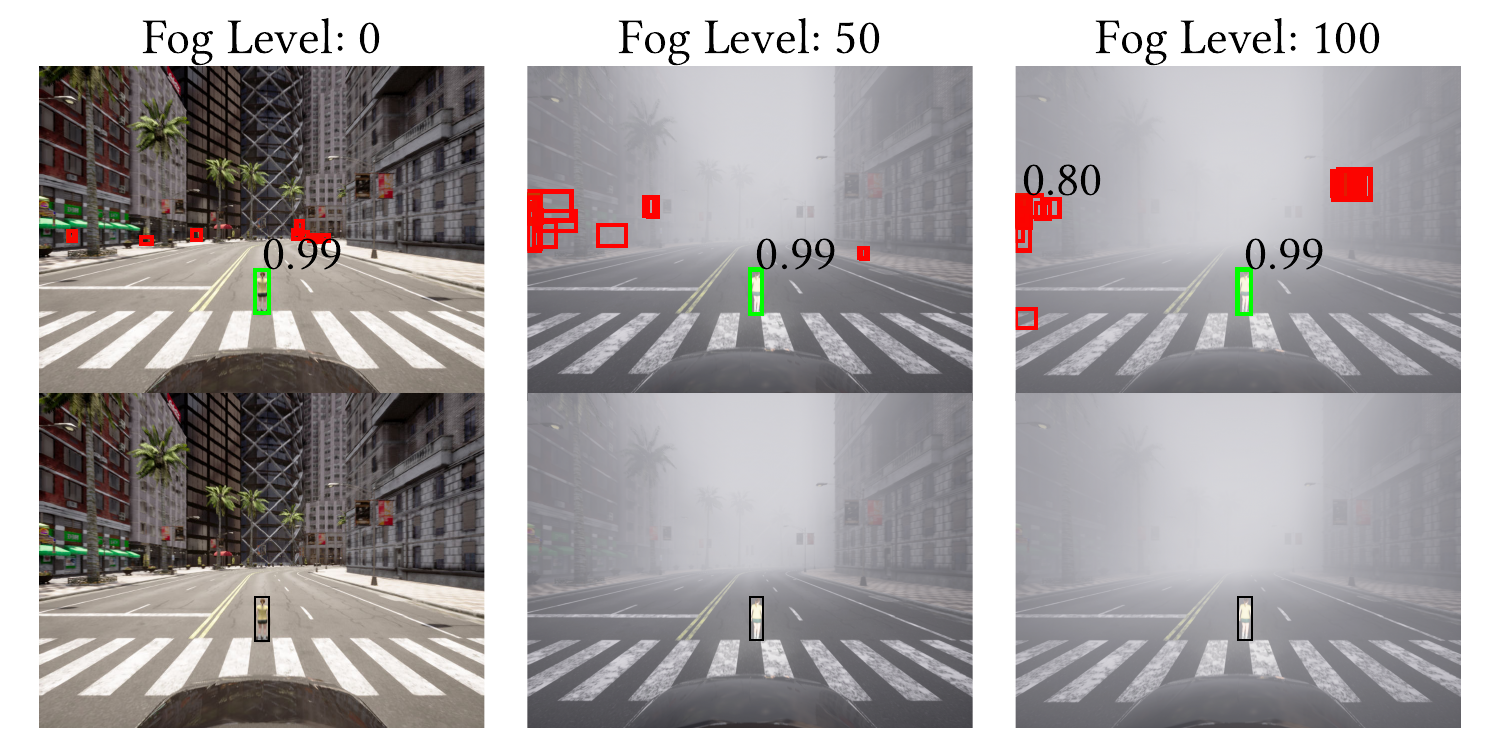}
    \caption{Black bounding boxes in the bottom plots indicate ground truth bounding boxes (i.e., there is only one pedestrian). Green indicates true positives and red indicates false positives. The confidence score is shown only when it is > 0.5. With the high levels of fog, it is possible to get false positives with confidences as high as 0.8.}
    \label{fig:plot-caleb}
 \end{minipage}%
\end{figure}

Object detectors are used to identify areas where objects such as pedestrians, vehicles, etc. are located in an image. Most deep learning-based object detectors, once trained, have to pass through two stages to predict these areas. In the first stage, even if there is only one object in an image, the object detector network provides multiple bounding box predictions as candidates. Subsequently, in the second stage, a careful analysis of all such predictions~\cite{neubeck2006efficient}, leads to the identification of the final areas containing objects (Figs.\ref{fig:preds}-~\ref{fig:plot-caleb}).

Each proposal prediction in the first stage comprises a tuple of bounding box coordinates, object class, and a corresponding confidence score. This confidence score measures the probability that a given bounding box belongs to the determined class. To compute the accuracy of the proposal predictions, they are sorted in ascending order based on the confidence scores and, are then assigned a ground truth bounding box if the intersection over union (IoU) surpasses a predefined threshold~\cite{girshick2015fast,carion2020endtoend}. Typically set to 0.5 or 0.75, this threshold determines the strictness of the criterion, with higher values indicating a stricter criterion. The various combinations of these can be summarized as follows:
\begin{itemize}[label={}, labelsep=0pt, leftmargin=0pt, itemsep=1em]
    \item {\bf True Positive} (TP): A predicted bounding box correctly represents a pedestrian by passing the IoU threshold. This is the only value we want it to be high.
    \item {\bf False Positive} (FP): A bounding box is predicted in a place where there are no pedestrians. In other terms, these are false alarms or hallucinations of pedestrians. 
    \item {\bf False Negative} (FN): In a place where there is a pedestrian, the object detector fails to place a bounding box. 
    \item {\bf True Negative} (TN): Since there can be finitely many bounding boxes that can be placed in areas where there are no pedestrians, TNs are not a valid metric in object detection. 
\end{itemize}

\subsection{Fairness Definitions for Object Detection}
\label{sec:fair_def}

In a classification setting, fairness is evaluated using metrics such as demographic parity~\cite{jiang2022generalized, barocas2023fairness}, equalized odds~\cite{hardt2016equality}, or equal opportunity~\cite{hardt2016equality}. Considering the high-stakes nature of the AV application, equalized odds and equal opportunity are better metrics because they focus on incorrect predictions as well. However, unlike in the classification setting, in object detection, 1) TNs are not defined and 2) an object detector provides many bounding box predictions with corresponding confidence scores. We consider a few metrics that aggregate all this information. Firstly, we consider two commonly used metrics in object detection. \\

\noindent {\bf Average Recall} (AR): This metric measures, out of all predicted bounding boxes, how many of them are correctly assigned to a ground truth bounding box.
\begin{equation}
    \mathrm{AR}= \frac{\mathrm{TP}}{\mathrm{TP} + \mathrm{FN}}
    \label{eq:ar}
\end{equation}
quantifies the object detector's effectiveness in detecting all pedestrians present in the image. A lower AR value indicates a higher number of FNs, which raises critical safety concerns. In the scenario where a person crosses in front of an AV equipped with a low AR object detector, there is a risk of the detector missing the person, potentially leading to fatal accidents. AR can be calculated at different IoU thresholds, such as AR@0.5, AR@0.75, or mAR (mean AR), which represents the average of all AR values from 0.5 to 0.95 with an increment of 0.05.\\

\noindent {\bf Average Precision} (AP): Out of all predicted proposal bounding boxes, how many of them are correctly assigned to ground truth bounding boxes is quantified using,
\begin{equation}
    \mathrm{AP}= \frac{\mathrm{TP}}{\mathrm{TP} + \mathrm{FP}}.
     \label{eq:ap}
\end{equation}
A low AP corresponds to a higher number of FPs, where the object detector hallucinates pedestrians who are not present in the scene. These unnecessary hallucinations can lead a vehicle to exhibit unnecessary caution, slowing down, or applying sudden brakes. Sudden braking, in particular, poses a risk of accidents. However, considering the consequences of lower AR versus lower AP, we put more emphasis on achieving a higher AR because there is a higher likelihood that a low AR value could result in direct collisions with pedestrians.

Since AR and AP do not account for confidence scores~\cite{Senanayake2024arxiv_unc}, they do not provide any information about how sensitive the missed detections and false predictions are. Therefore, we define two new metrics.\\ 

\noindent {\bf Average True Positive Confidence} (ATPC): 

Aggregating confidences of TPs, we define,
\begin{equation}
    \mathrm{ATPC} = \frac{1}{N_{\mathrm{TP}}}\sum p(\hat{y}=1|x,\mathrm{class=person})
     \label{eq:atpc}
\end{equation}
where $N_{\mathrm{TP}}$ is the total number of TPs for image $x$ when the ground truth $\mathrm{class = person} \in \mathcal{C}$. All the object classes, $\mathcal{C}$, the object detector can identify, for instance, include persons, trees, cars, cats, and $\phi$, not a class. Even if a predicted proposal bounding box is classified as a TP because it passed the IoU threshold, it might be detected as a different object, when the vehicle is moving, especially in challenging weather. In the Arizona accident, according to~\cite{Madrigal.2018}, the person was misdetected ``as an unknown object, as a vehicle, and then as a bicycle with varying expectations of future travel path,'' resulting in a fatal accident. Since a higher TP is not useful if the confidences are low, object detectors need to have a higher ATPC as well.\\

\noindent {\bf Average False Positive Confidence} (AFPC):
Aggregating confidences of FPs, we define,
\begin{equation}
    \mathrm{AFPC} = \frac{1}{N_{\mathrm{FP}}}\sum p(\hat{y}=1|x,\mathrm{class}={\phi}). 
     \label{eq:afpc}
\end{equation}
where $N_{\mathrm{FP}}$ is the total number of TPs. In the presence of hallucinations, lower confidence scores are desirable as they enable the vehicle to effectively filter them out as false alarms.

While improving the object detectors based on these four metrics is important, to guarantee fairness, we also need to make sure the metrics across different groups are equal. For protected attributes $a \in \mathcal{A}$, the groups $g$ can be a single protected group $g_a$ such as female or an intersectional group $g_{a_1 \times a_2 \times ... a_m} = g_{a_1} \cap g_{a_2} ... \cap g_{a_m}$ such as white-female-child. For weather conditions $w \in \mathcal{W}$, $g$ can also be a group that consists of protected attributes and weather conditions $g_{(a_1 \times a_2 \times ... a_m) \times (w_1 \times w_2 \times ... w_n)} = g_{a_1 \times a_2 \times ... a_m} \cap g_{w_1 \times w_2 \times ... w_n}$ such as white-female-child under dark-foggy weather.

For groups $g$ of interest, equal opportunity is classically defined as,
$P(\hat{y}=1|\mathcal{A} \in g_i, y=1) = P(\hat{y} = 1 | \mathcal{A} \in g_j, y = 1), \forall i, j$ group indices. In order for an object detector to be fair, considering all the metrics defined in  (\ref{eq:ar})-(\ref{eq:afpc}), for all $i, j$ pairs, the following equalities should hold:

\begin{subequations}
\begin{gather}
    AR_{g_i} = AR_{g_j} \label{eq:subeq_a}\\
    AP_{g_i} = AP_{g_j} \label{eq:subeq_b}\\
    ATPC_{g_i} = ATPC_{g_j} \label{eq:subeq_c} \\
    AFPC_{g_i} = AFPC_{g_j} \label{eq:subeq_d}
\end{gather}
\end{subequations}

\subsection{Fairness Comparisons}
\label{sec:fair-mets}

For any metric $s$ defined in equations \ref{eq:subeq_a}-\ref{eq:subeq_d} and $i,j$ group indices,

\noindent {\bf Worst-case difference}: This metric measures the largest possible disparity between two demographic groups.
\begin{equation}
    {\Delta}_\mathrm{worst}s = \max \Delta_{g_i-g_j} = \max | s_{g_i} - s_{g_j} |, \forall i,j
\end{equation}

\noindent {\bf Best-case difference}: This metric measures the smallest possible disparity between two demographic groups.
\begin{equation}
    {\Delta}_\mathrm{best}s = \min \Delta_{g_i-g_j} = \max | s_{g_i} - s_{g_j} |, \forall i,j
\end{equation}

\noindent {\bf Wasserstein-2 metric}: This metric measures maximum Wasserstein-2 metric~\cite{ramdas2017wasserstein} of all groups of interest. In other words, it indicates how similar the worst-case metrics are. A lower value indicates less disparity. 
% \begin{equation}
%     W_s = \max_{i,j} \Large(\min_{P \in \mathbb{R}}\sum_{i^\prime,j^\prime} P_{i^\prime j^\prime} \| s_g_{i^\prime} - s_g_{j^\prime} \|_2^2 \Large)^\frac{1}{2}, \forall i,j
% \end{equation}

\begin{equation}
    W_s = \max_{i,j} \left( \min_{P \in \mathbb{R}} \sum_{i', j'} P_{i' j'} \| s_{gi'} - s_{gj'} \|_2^2 \right)^{\frac{1}{2}}, \forall i,j
\end{equation}

\section{Models and Datasets}

\subsection{Transformer-Based Object Detection}
We used DETR~\cite{carion2020endtoend} with a ResNet-50 backbone, one of the best-performing transformer-based object detection models, as the pedestrian detector. It is a 41 million parameter object detection model trained on the MS COCO dataset~\cite{lin2015microsoft} and is capable of detecting 80 object types. As we are interested in the effect on pedestrians, we only focus on the ``person'' object type. While these training images contain 262,465 people with diverse attributes in diverse backgrounds, the exact composition is not known.

\subsection{FACET Dataset}

\begin{figure}
    \centering
    \includegraphics[width=0.5\linewidth]{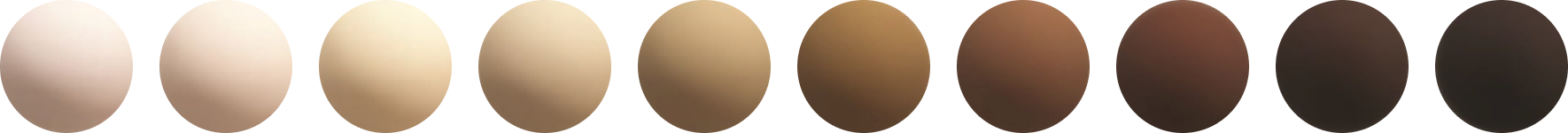}
    \caption{Monk Skin Tone (MST)~\cite{Monk_2019} scale where MST=1 is the lightest skin tone and MST=10 is the darkest skin tone}
    \label{fig:mst}
\end{figure}

FACET~\cite{gustafson2023facet} is the latest publicly available fairness evaluation dataset containing 32,000 images that are meticulously annotated by expert reviewers to find the disparities in common computer vision tasks such as image classification, object detection, and segmentation. The dataset contains 50,000 person annotations with not only manually drawn bounding boxes but also person-related attributes such as perceived skin tone, hair type, and person class. In our work of finding the performance disparities of object detection models, we are mainly interested in the perceived skin tone and the lighting condition annotations of the dataset. \\
\textbf{Perceived skin tone}: The FACET dataset utilizes the Monk Skin Tone (MST) scale shown in Fig.~\ref{fig:mst}. The MST scale is an inclusive skin tone scale that includes 10 different skin tones where MST=1 being the lightest skin tone and MST=10 being the darkest skin tone of the scale. In FACET, the skin tone is annotated as a spectrum rather than a single tone due to the difficulty of pinpointing an exact skin tone, especially when considering the effect of lighting conditions. Since a single annotation can have multiple skin tones annotated, we processed our data following~\cite{gustafson2023facet}.\\
\textbf{Lighting condition}: Each annotation is categorized into overexposed, underexposed, dimly-lit, and well-lit based on the perceived lighting condition. In our work, we do not focus on overexposed and underexposed lighting conditions because they refer to camera settings rather than the true ambient light level. 

\subsection{Carla Simulation}

Carla~\cite{pmlr-v78-dosovitskiy17a} is an open-source high-fidelity driving simulator designed to create near-real-world environments for experiments on vehicles. Since Carla is powered with various customization options such as the layouts of the town, weather conditions, pedestrian types, and vehicles as well as a variety of sensors for comprehensive data collection, it makes the best choice for creating a custom evaluation dataset.

For data collection, we utilized an autopilot-enabled car with both an RGB camera and a camera that captured semantic segmented ground truth directly from the simulation every 5 seconds of the simulation to create a dataset for selected types of pedestrians in a set weather condition with each experiment roughly yielding 1000 images. We consider 10 different person types varying in gender, body type, and skin tone. While we mostly present our analysis on foggy weather conditions, we also tested on cloudy and rainy weather conditions. The intensity of each weather condition is scaled from 0\% to 100\% with 0\% being no adverse weather and 100\% being the most extreme weather. Data was collected at 0\%, 25\%, 50\%, 75\%, and 100\% intensity levels. 

We noticed that the benefits of extracting bounding boxes from the semantic segmented images using the simulator's rendering engine over Carla's built-in bounding box extraction tools are twofold. First, this approach not only accelerates the data collection process but also preserves the accuracy even at high data sampling speeds. Second, and most importantly, the semantic segmentation method inherently filters out objects that are not in the direct line of sight of the ego vehicle, such as pedestrians behind buildings — a feat that Carla's default tool does not achieve without additional processing steps. To maintain the consistency across different experiments, random seeds were used for all the actors of the environment expecting the outputs would be deterministic, although we still experienced negligible levels of randomness. One of the main drawback of using the current version of Carla stemmed from its limited selection of pedestrian models from a catalog of 50 pedestrian variants which also did not allow for straightforward customization.

\begin{figure}
    \centering
    \includegraphics[width=1\linewidth]{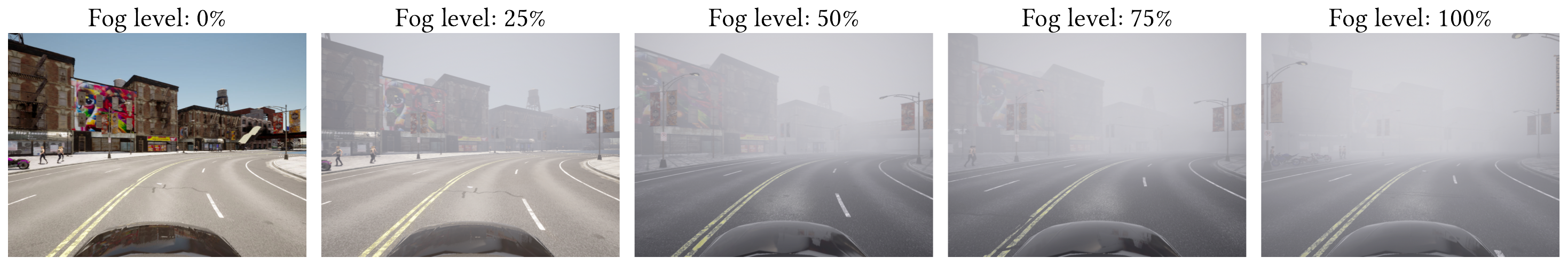}
    \caption{Carla simulation sample image across fog intensities of 0\%, 25\%, 50\%, 75\%, and 100\%. The visibility of the road incrementally reduces as the fog intensity increases.}
    \label{fig:fog-scale}
\end{figure}

\section{Experiments}

We conduct experiments to answer the following questions:
\begin{itemize}[label={}, labelsep=0pt, leftmargin=0pt]
    \item Q1. Is object detection sensitive to protected attributes $\mathcal{A}=\{\mathrm{skin tone, gender, age}\}$?
    \item Q2. Does object detection performance degrade with weather conditions $\mathcal{W}=\{\text{ambient light level, fog, rain}\}$ for protected attributes $\mathcal{A}$?
\end{itemize}

\subsection{Fairness Analysis on the FACET dataset}

\begin{figure}
    \centering
    \includegraphics[width=1\linewidth]{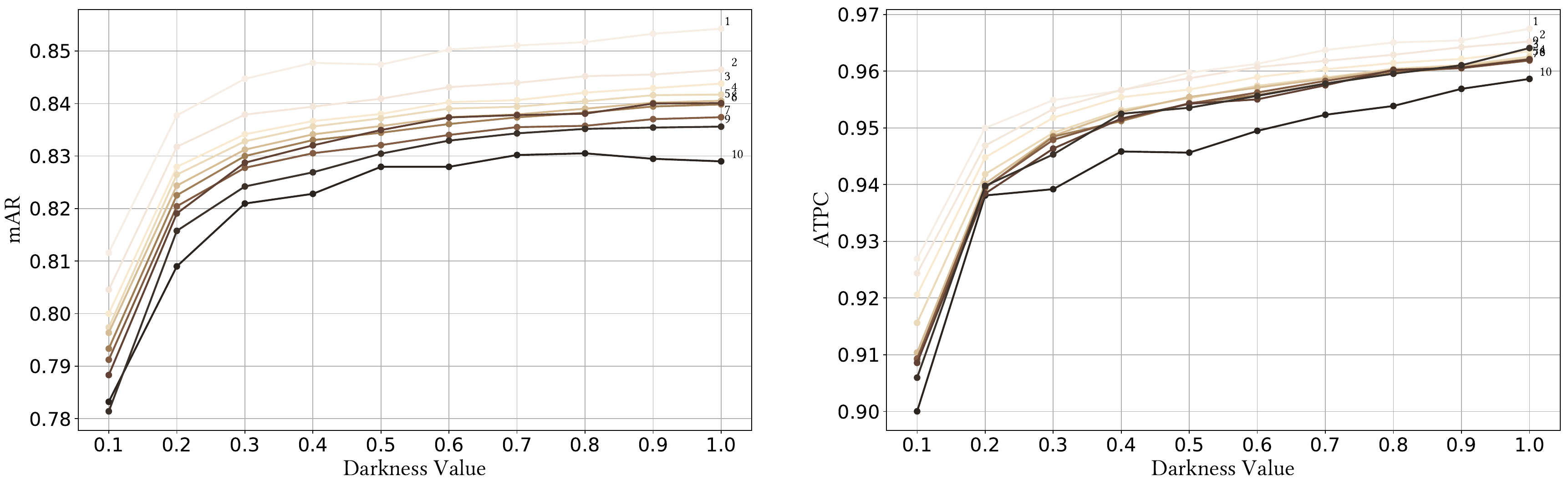}
    \caption{Performance disparities of ResNet50-backbone DETR model on Monk Skin Tone scale on FACET dataset. The metrics mAR and ATPC shows the model is more capable of identifying lighter skin tone people more confidently than darker skin tone people. For any skin tone, the model's performance drops with the ambient darkness (0=dark).}
    \label{fig:plot-darkvsskin}
\end{figure}

\subsubsection{The effect of skin tone vs. artificial darkness (Fig.~\ref{fig:plot-darkvsskin})}
\label{sec:facet_skintone}

\begin{wrapfigure}{r}{0.5\textwidth} 
  \centering
  \includegraphics[width=0.48\textwidth]{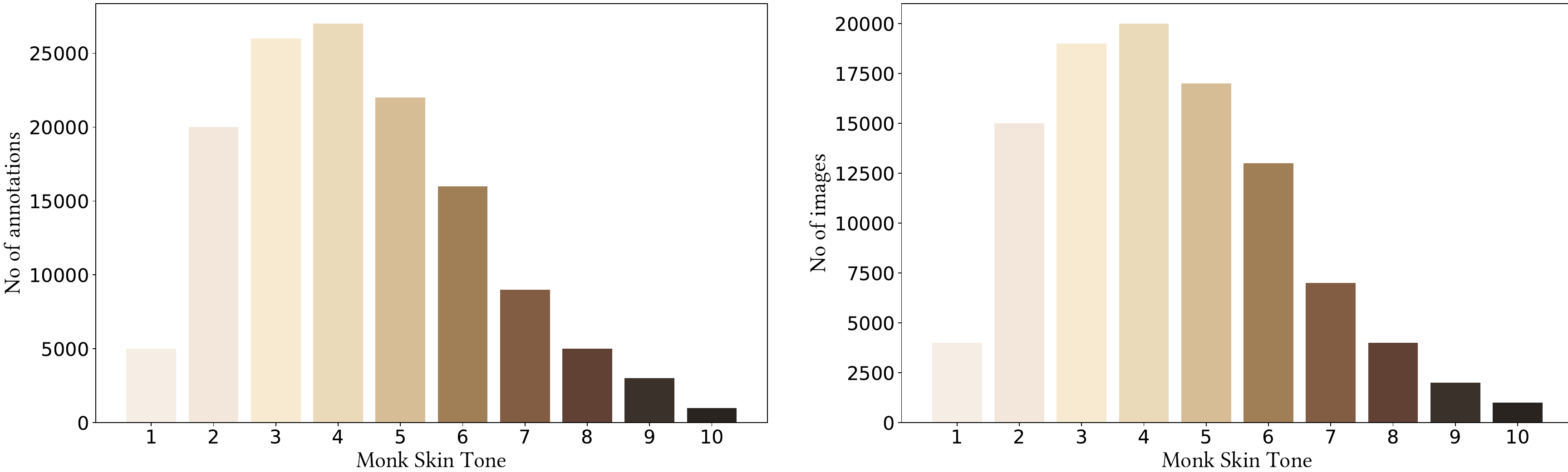} 
  \caption{Histogram of the images and annotation with Monk Skin Tone scale for FACET dataset. Lighter skin tone annotations are more prominent in the dataset compared to the darker skin tone annotations.}
  \label{fig:plot-histo}
  \vspace{-15pt}
\end{wrapfigure}
% \begin{figure}
%     \centering
%     \includegraphics[width=1\linewidth]{images/histogram_2.pdf}
%     \caption{Histogram of the images and annotation with Monk Skin Tone scale for FACET dataset. Lighter skin tone annotations are more prominent in the dataset compared to the darker skin tone annotations.}
%     \label{fig:plot-histo}
% \end{figure}

\begin{figure}
    \centering
    \includegraphics[width=1\linewidth]{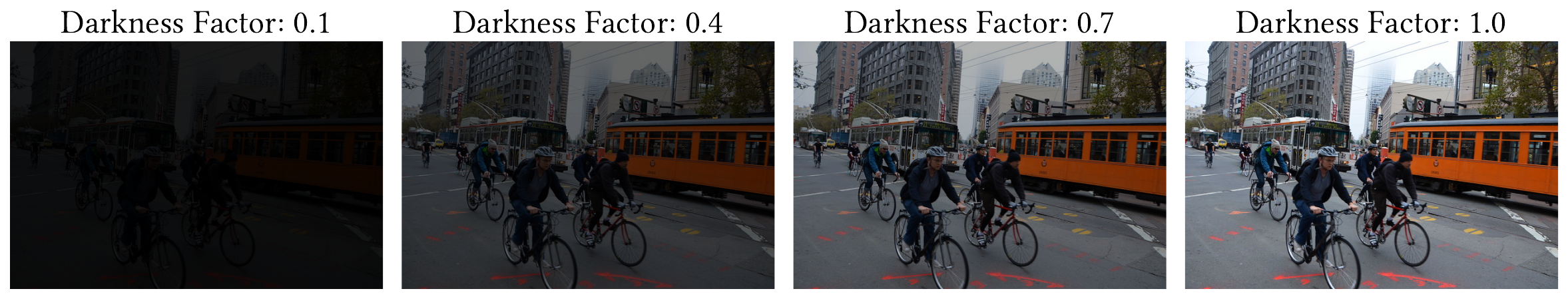}
    \caption{FACET sample image across processed darkness levels 0.1, 0.4, 0.7, and 1.0 with 1.0 reflecting original darkness level and 0.0 representing total darkness. The darkness achieved through image processing techniques are intended to mimic natural lighting conditions.}
    \label{fig:darkness-scale}
\end{figure}

We evaluate the DETR model by following the evaluation procedure in FACET~\cite{gustafson2023facet}. Additionally, to simulate the ambient darkness in FACET images, we multiplied pixel values with darkness factors ranging from 0 to 1 with an increment of 0.1~\cite{gonzalez2009digital}. The darkness factor 0 indicates a completely dark image and 1 indicates the original image (Fig.~\ref{fig:darkness-scale}). We focused on the model performance for person detection for Monk scale skin tones under artificially introduced 10 levels of darkness factors to assess how the model biases vary with respect to the weather conditions.

As shown in Fig.~\ref{fig:plot-darkvsskin}, for all skin tones, the darker the environment is, the lower the model performance is. Further, under all ambient darkness values, the model has the best performance for the fairest skin tone and the poorest performance for the darkest skin tone. In general, fairer skin tones outperform darker skin tones.

\begin{figure}
    \centering
    \includegraphics[width=1\linewidth]{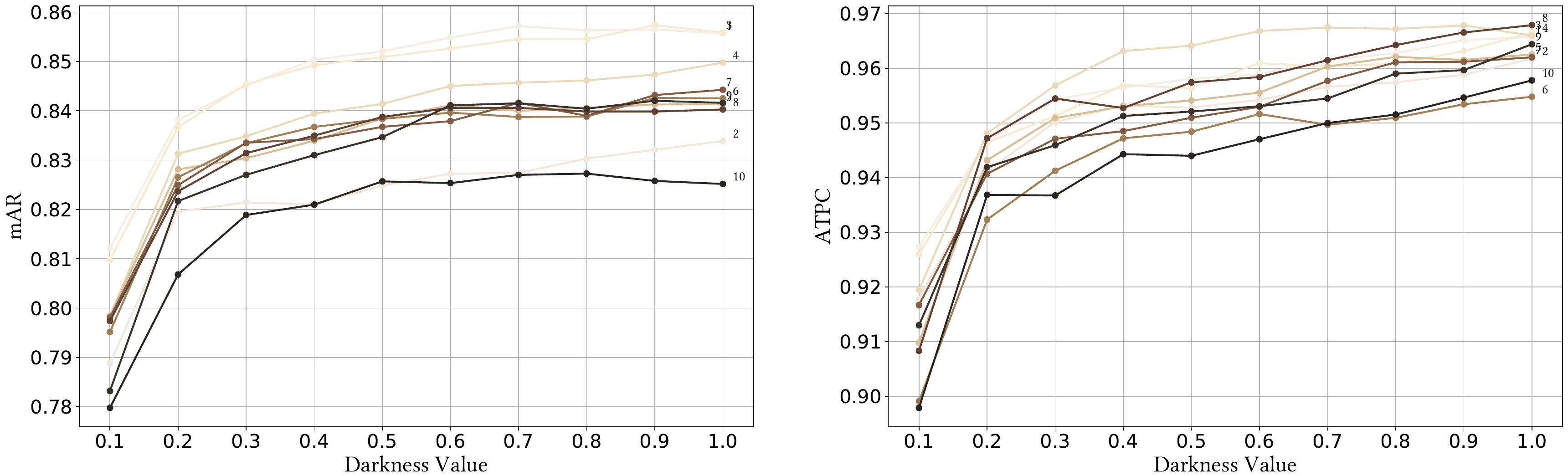}
    \caption{Evaluation of skin color and darkness on FACET dataset with Monk Skin Tone scale using same sample size. The results indicate a general advantage for lighter skin tones in the mAR metric but no significant disparity in ATPC metric. The disparity trends can vary even for randomly selected subsets and the full dataset. }
    \label{fig:plot-darkvsskin1000}
\end{figure}

\subsubsection{The effect of skin tone vs. artificial ambient darkness on an equal sample size (Fig.~\ref{fig:plot-darkvsskin1000})}
\label{sec:facet_skintone_samesize}

As shown in the histograms of Fig.~\ref{fig:plot-histo}, the number of skin tone annotations for darker skins is relatively small. Therefore, we evaluated the model on a randomly selected 1000 images for each Monk scale skin tone. Although the model performance deteriorates with ambient darkness (Fig.~\ref{fig:plot-darkvsskin1000}), people with fairer skin tones seem to be better advantaged. However, the APTC values have a higher variance compared to experiments in Section~\ref{sec:facet_skintone}, indicating the sensitivity of probabilities to the sample size. 

\begin{figure}
    \centering
    \includegraphics[width=1\linewidth]{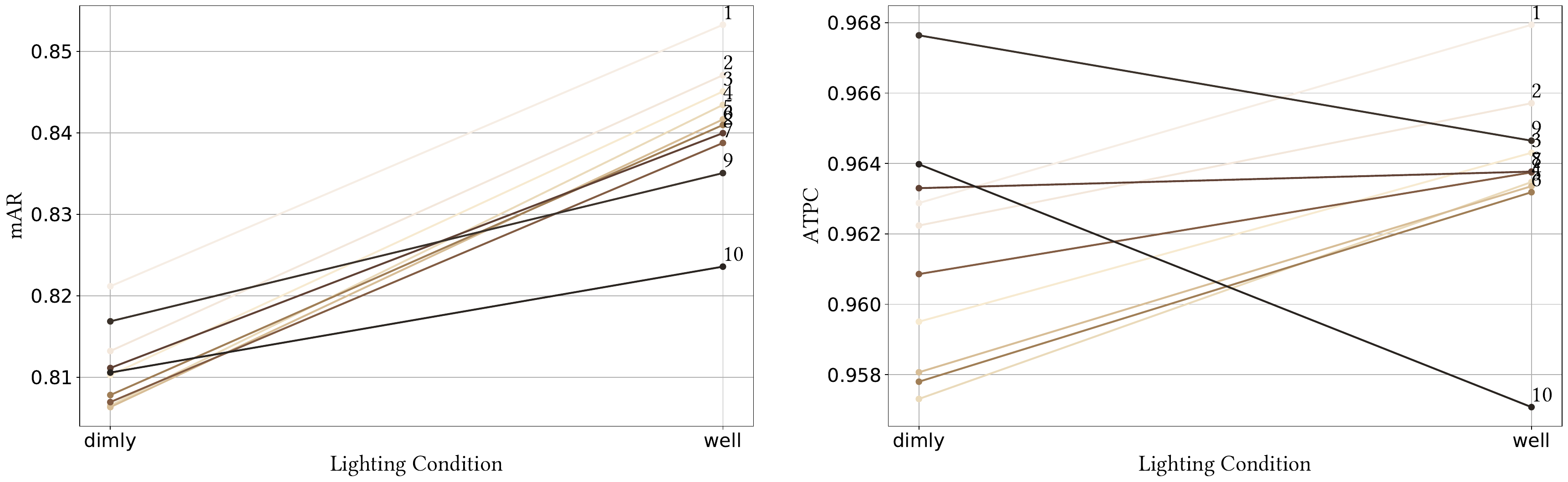}
    \caption{Analysis on the annotated lighting conditions, "well-lit" and "dimly-lit", and the skin tone. While the disparity for skin tones in the dimly-lit is not significant, lighter skin tones stands a better way of getting identified in well-lit conditions. }
    \label{fig:plot-light}
\end{figure}

\subsubsection{The effect of skin tones vs. annotated lighting conditions (Fig.~\ref{fig:plot-light})}

The FACET annotations include the perceived ambient light condition categorized as ``well-lit,'' ``dimly-lit,'' ``overexposed,'' and ``underexposed.'' Unlike our prior experiment that assessed the model performance across various skin tones under darkness levels modified through image processing methods, this study focuses only on the annotated natural lighting conditions. Therefore, we exclusively consider the annotations categorized under ``well-lit'' and ``dimly-lit'' since they reflect the lighting condition of the environment while ``overexposed'' and ``underexposed'' indicate camera exposure settings that fall out of the scope of our work.

As shown in Fig.~\ref{fig:plot-light}, while in dimly-lit conditions, the model performs on par for both darker and lighter skin tones, especially in terms of the ATPC metric. However, when the lighting is optimal, individuals with lighter skin tones stand a better chance of being recognized by the object detector and tend to be recognized with a higher confidence score on average.

\begin{figure}
  \begin{subfigure}[b]{1\textwidth}
    \centering
    \includegraphics[width=1\linewidth]{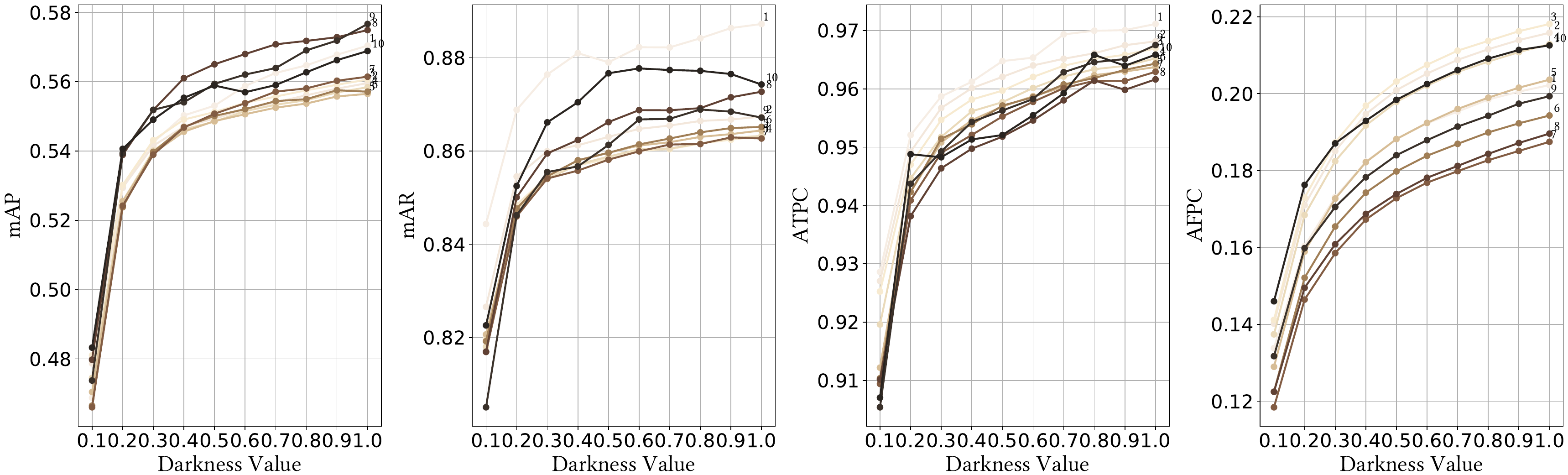}
    \caption*{(a) Analysis on filtered skin tone annotations with ambient darkness levels (0=dark). }
    \label{fig:plot-filtered}
  \end{subfigure}%
  \hfill
  \begin{subfigure}[b]{1\textwidth}
    \centering
    \includegraphics[width=1\linewidth]{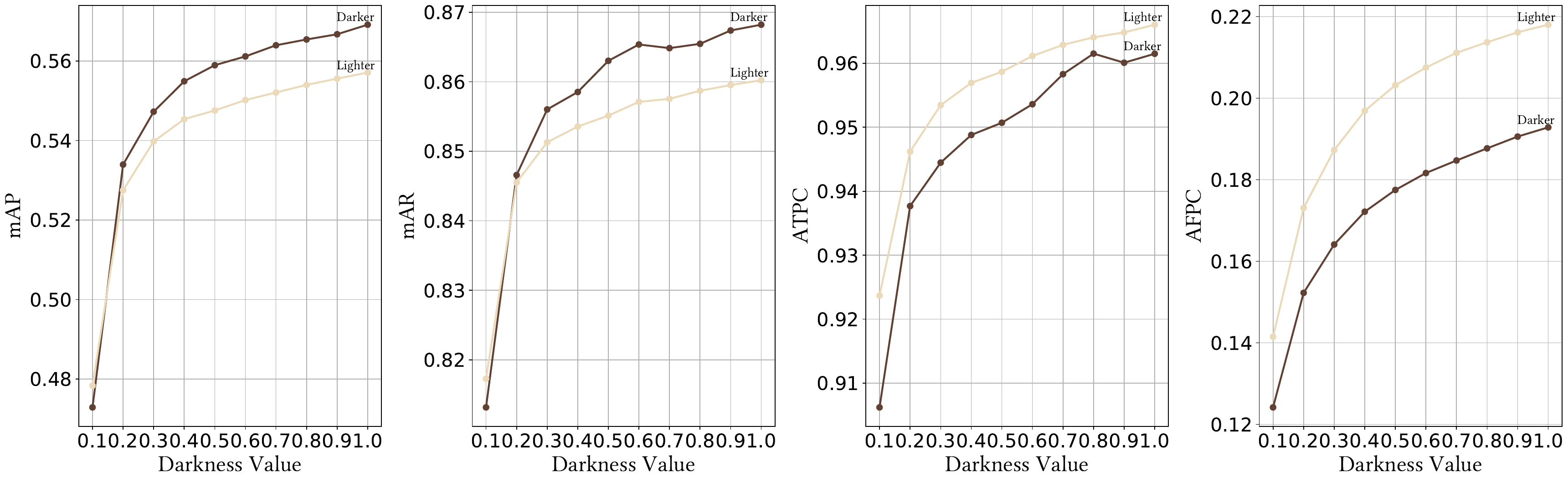}
   \caption*{(b)  Analysis on filtered and grouped skin tone annotations with ambient darkness levels (0=dark).}
    \label{fig:plot-filtered-darkfair}
  \end{subfigure}%
  \caption{Analysis on filtered skin tone annotations and darkness levels shows the model outperforms on darker skin tones compared to lighter skin tones in all the metrics except for ATPC. For a clearer comparison of the model performance disparities for filtered dataset on skin tone, the lighter skin tone, MST=(1, 2, 3), and the darker skin tones, MST=(8, 9, 10), are grouped together and evaluated. The trend of (a) is more visible in the grouped darker skin and lighter skin plot (b).}
  \label{fig:filtered_ab}
\end{figure}

\subsubsection{The effect of filtered skin tone annotations vs. ambient darkness levels (Fig.~\ref{fig:filtered_ab})}
\label{sec:facet_skintone_filtered}

Since the concept of FPs become ambiguous when there are multiple people in an image~\cite{gustafson2023facet}, in Section~\ref{sec:facet_skintone}, we only focused on metrics that do not take into account FPs. The demographic attribute of a predicted bounding box is not clear when the evaluated image contains multiple annotated people with different attributes. Therefore, we curate the FACET dataset to obtain a subset of annotations pooled by excluding the images that have multiple ground truths with different Monk skin tone values. For example, the filtered MST=1 evaluation dataset contains only the images that all the annotations have the skin tone label of MST=1. This curation allows us to use AP and AFPC for model performance evaluation.

As illustrated in Fig.~\ref{fig:filtered_ab}a, the model outperforms for darker skin tones across all the metrics except for ATPC, which shows that the model is more confident when correctly identifying a person with a fairer skin tone. For a clearer comparison between the lighter and darker skin tones, we grouped MST=(1, 2, 3) as the lighter skin tone and MST=(8, 9, 10) as the darker skin tone in Fig.~\ref{fig:filtered_ab}b. Results in both plots contradict the results in Sections~\ref{sec:facet_skintone}-\ref{sec:facet_skintone_samesize}. This contradiction indicates that if there are people belonging to different demographic groups, the object detector performance varies. This discrepancy is potentially because of the number of predicted bounding boxes and their associated confidence levels distributed unequally among people with different skin tones. To eliminate these confounding factors, it is ideal to use a simulator.

\subsection{Fairness Evaluation using Carla}

Each Carla experiment has a unique pedestrian type and unique weather type or weather severity. This approach of using only one type of pedestrian for the whole town per experiment helps us evaluate the object detector exhaustively for all the metrics including those involving FPs. Additionally, the use of one type of weather per experiment, allows us to evaluate what specifically causes the object detector to fail.

\begin{figure}
  \begin{subfigure}[b]{1\textwidth}
    \centering
    \includegraphics[width=1\linewidth]{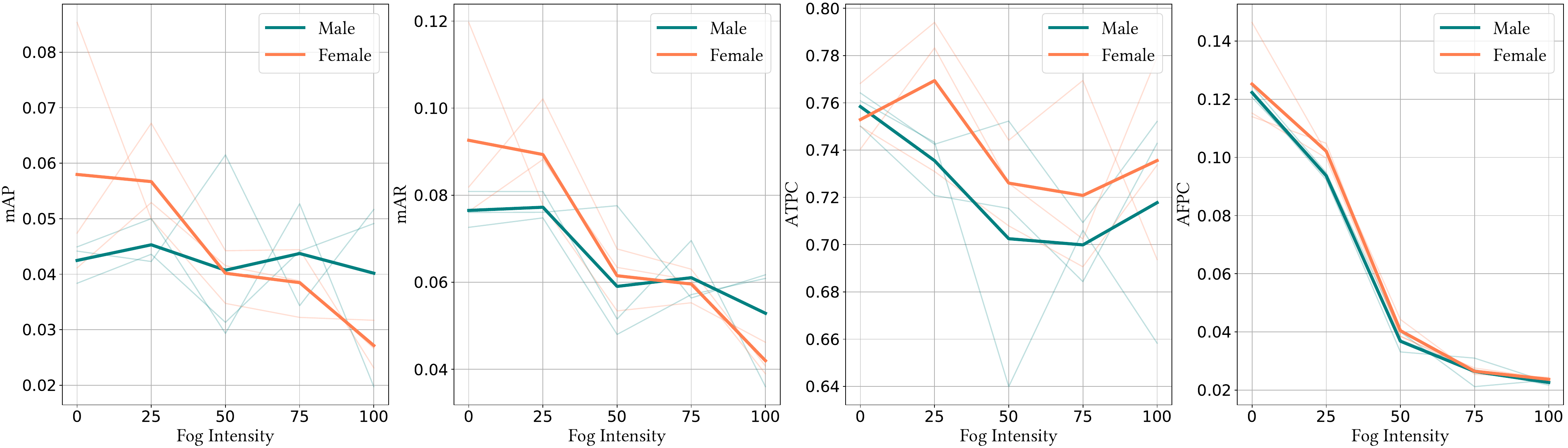}
    \caption*{(a) Gender}
    \label{fig:sub_gender}
  \end{subfigure}%
  \hfill
  \begin{subfigure}[b]{1\textwidth}
    \centering
    \includegraphics[width=1\linewidth]{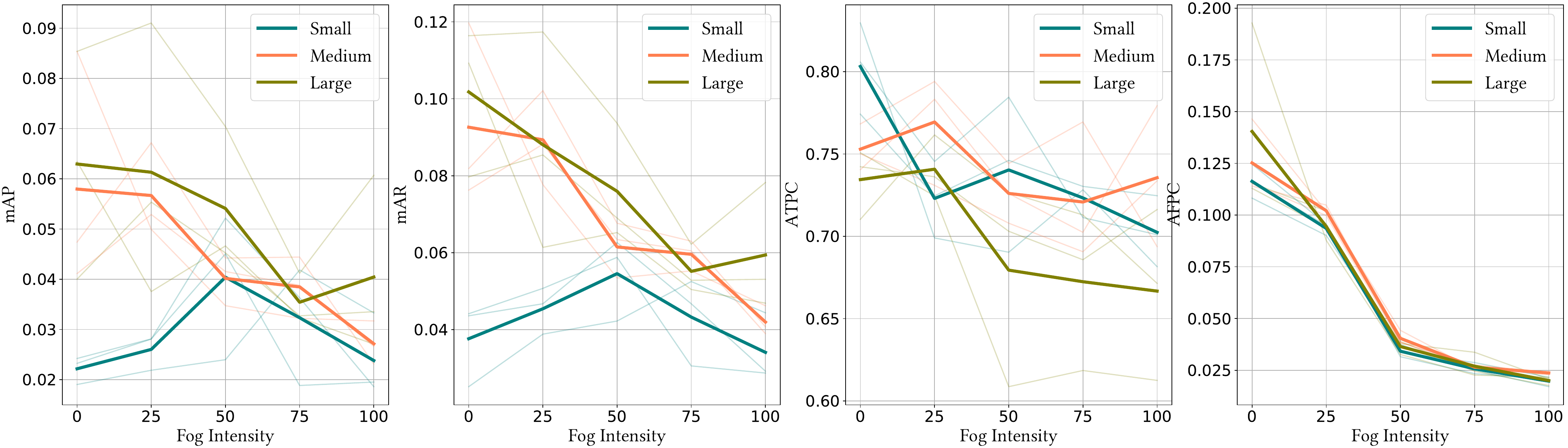}
    \caption*{(b) Body size}
    \label{fig:sub_bodysize}
  \end{subfigure}%
  \hfill
  \begin{subfigure}[b]{1\textwidth}
    \centering
    \includegraphics[width=1\linewidth]{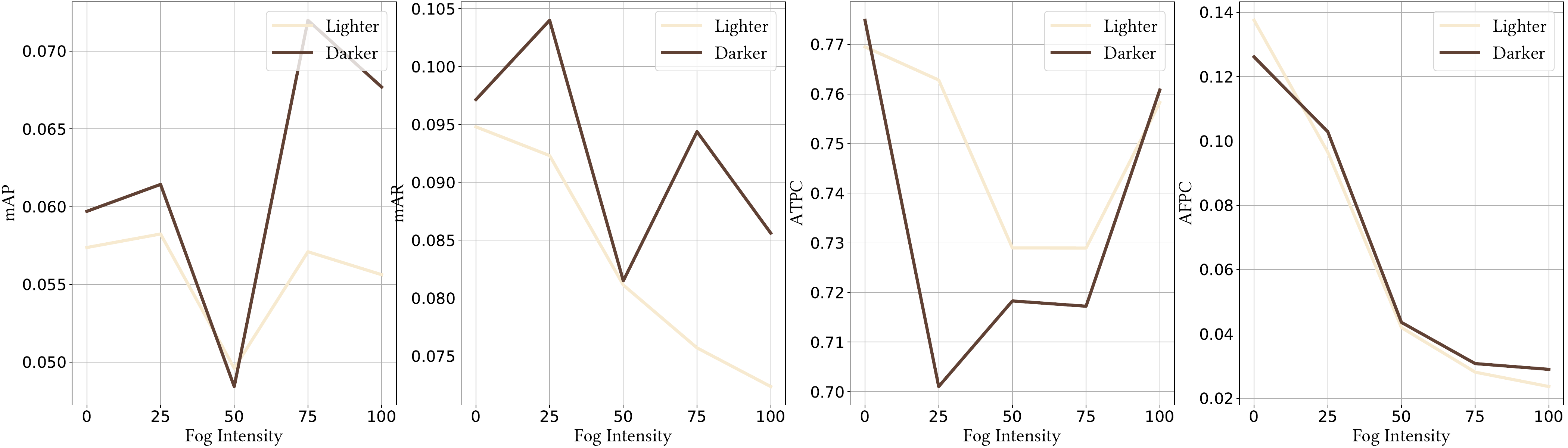}
    \caption*{(c) Skin tone}
    \label{fig:sub_skin}
  \end{subfigure}
  \caption{The disparity analysis of the performance of the object detector for Carla experiments based on gender, skin tone and the body sizes are shown. Even though gender and skin tone do not affect much, small body size (children) is detected poorly. Each category of gender and body size plots use three types of pedestrains and the bold line represents the average line for each group. The skin tone plots use only one pedestrian type for each category due to the limited choices in the Carla catalogue of pedestrians. }
  \label{fig:3x4plot}
\end{figure}

\subsubsection{The effect of protected attributes vs. weather conditions (Fig.~\ref{fig:3x4plot})}
\label{sec:trends}

Almost all the pre-defined pedestrian types of Carla are fully clothed and the only exposed areas of the skin are the face and hands. Additionally, to make the evaluation as fair as possible, we insisted on comparing the performance of the object detector with two pedestrian types that wear similar clothes. As shown in Fig.~\ref{fig:3x4plot}, performance, in general degrades with fog intensity. However, considering the magnitude of each metric, gender and skin tone do not seem to affect much while the small body sizes (i.e.. children) are always affected. To assess these differences quantitatively, by considering the metrics defined in Section~\ref{sec:fair-mets}, Table~\ref{tab:mid-scale-table} was created. As a general trend, we can observe that the disparity between demographic groups reduces with the increasing fog level. 

\begin{figure}
  \begin{minipage}[b]{0.5\textwidth}
    \centering
    \includegraphics[width=\textwidth]{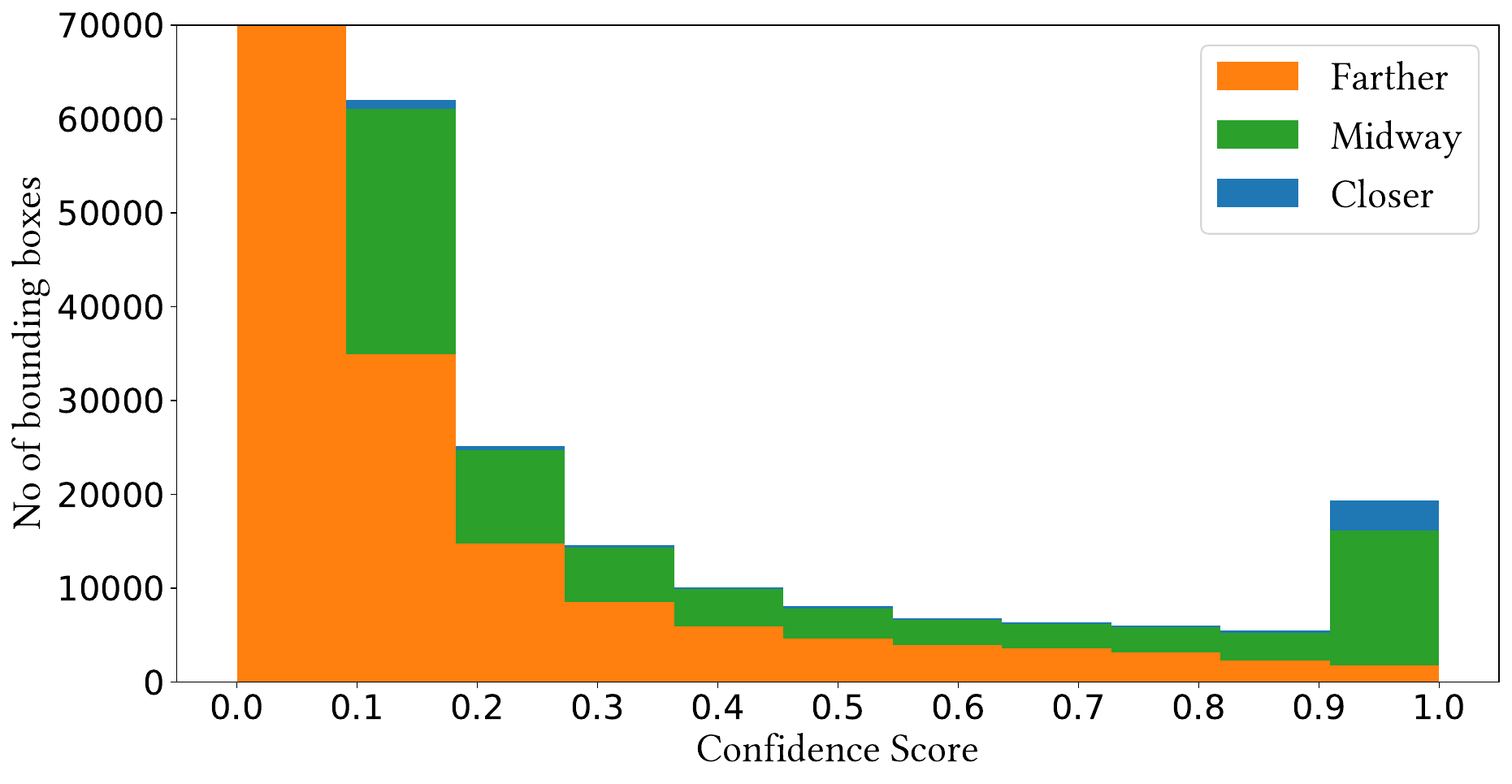}
    \caption{Aggregated confidence distribution of the predicted bounding boxes of Carla experiments. The predicted bounding boxes are categorized into ``farther,'' ``midway,'' and ``closer'' based on their corresponding area sizes - large, medium and small as per ~\cite{lin2015microsoft}. For the bounding boxes predicted in the ``farther'' category, the model predominantly exhibits lower confidence levels, indicating a potential correlation between distance and predictive confidence. }
    \label{fig:plot-jamaica-distrib}
  \end{minipage}
  \begin{minipage}[b]{0.45\textwidth}
    \centering
    \includegraphics[width=0.7\textwidth]{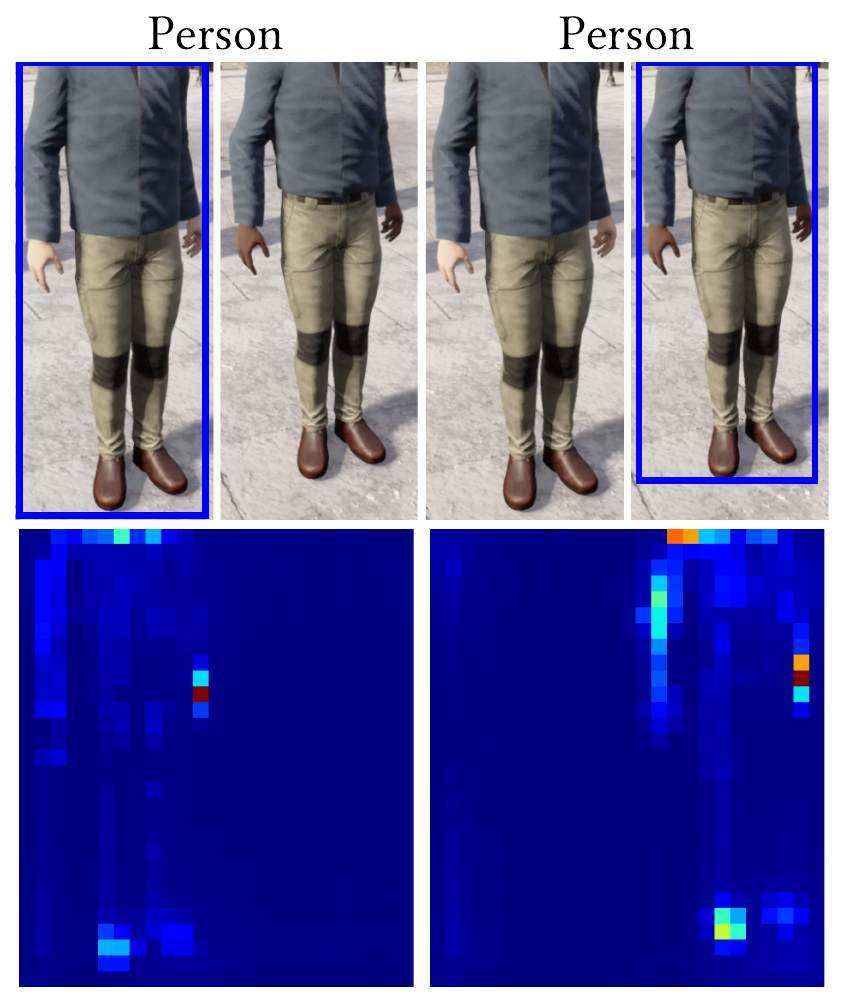}
    \caption{Top row images show two people in the same clothing but with different skin tones (MST=2 and MST=8) and heads occluded. Blue indicates detected bounding boxes. The bottom row shows the attention maps of the final decoder layer of the DETR model. The attention heat map indicates the whole body is utilized rather than one body part.}
    \label{fig:attention_maps}
 \end{minipage}%
\end{figure}

\begin{table}
  \centering
    \caption{Application of disparity metrics mentioned in Section~\ref{sec:fair-mets} to mAR in gender and body size experiments conducted using Carla simulated environments. The skin tone evaluation is neglected due to the utilization of only a single person ID per skin tone category. ${\Delta}_\mathrm{worst}mAR$ reports the maximum observed mAR disparity and ${\Delta}_\mathrm{best}mAR$ reports the minimum. $W_{mAR}$ is used to calculate the Wasserstein distance between mAR distributions of each group. Both worst-case disparity and Wasserstein distances show a decrease in the disparity with the fog intensity level. }
    \begin{tabular}{lSSSSSS}
    \toprule
     & \multicolumn{3}{c}{Gender} & \multicolumn{3}{c}{Body size}\\
    \cmidrule(r){2-4}\cmidrule(l){5-7}
    Fog levels  & ${\Delta}_\mathrm{worst}mAR$ & ${\Delta}_\mathrm{best}mAR$ & $W_{mAR}$ & ${\Delta}_\mathrm{worst}mAR$  & ${\Delta}_\mathrm{best}mAR$ & $W_{mAR}$\\
    \midrule
    0\%  & 4.71 & 0.02 & 0.05 & 9.45 & 0.22 & 0.42 \\
    \midrule
    25\% &  2.73 & 0.16 & 0.02 & 7.85 & 0.28 & 0.21 \\
    \midrule
    50\% &  2.42 & 0.19 & 0.01 & 5.14 & 0.08 & 0.05 \\
    \midrule
    75\% &  1.43 & 0.11 & 0.00 & 3.24 & 0.03 & 0.03 \\
    \midrule
    100\% &  2.27 & 0.29 & 0.02 & 4.95 & 0.07 & 0.07 \\
    \bottomrule
    \end{tabular}
  \label{tab:mid-scale-table}
\end{table}

Since we did not observe a significant disparity based on gender or skin tone, we inspected the final decoder layer attention maps of the transformer model (Fig~\ref{fig:attention_maps}). Based on visual inspection, we speculate that the transformer looks at the person as a whole rather than the exposed skin patches because the attention modules are designed to look globally rather than the nearby pixels. This leads to the question of how clothing affects detection scores, although we do not have data to assess the same person under different clothing types and colors. All these deeper questions further highlights the importance of developing more sophisticated simulators.

\begin{figure}
    \centering
    \includegraphics[width=1\linewidth]{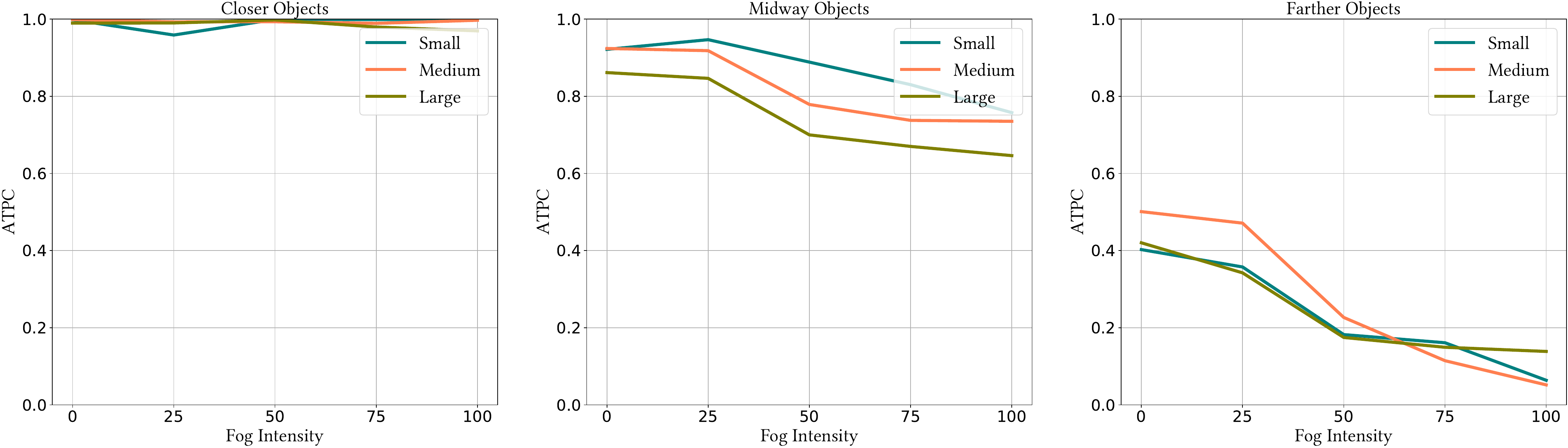}
    \caption{ATPC metric values of Fig.~\ref{fig:3x4plot}b is divided into three plots based on the distance of the annotations to the ego vehicle. The ATPC metric decreases with both the distance and fog intensity. }
    \label{fig:plot-atpc-distance}
\end{figure}

\subsubsection{The effect of object distance (Fig.~\ref{fig:plot-jamaica-distrib}-~\ref{fig:plot-atpc-distance})}
\label{sec:carla_distance}

We grouped the confidence scores of all pedestrians used in the Carla dataset using their distance from the vehicle. As illustrated in Fig.~\ref{fig:plot-jamaica-distrib}, most farther pedestrians have a lower confidence score. Therefore, assuming that distance plays a role in fairness, metrics were recalculated for the most significant disparity, the body size. As shown in Fig~\ref{fig:plot-atpc-distance}, farther pedestrians not only have a lower ATPC but also their ATPC values decrease with higher fog levels, indicating that. Further, the body size disparity does not exist for pedestrians closer to the vehicle.

\section{Discussions, Recommendations, and Limitations}

Experiments revealed several confounding factors in SOTA transformer-based object detectors. Our findings can be summarizes as follows:
\begin{itemize}
    \item Table~\ref{tab:mid-scale-table} shows that the disparity values for body size and gender reduces with the fog intensity.
    \item Fig.~\ref{fig:3x4plot} shows how the pedestrians with larger body sizes or darker skins were accurately identified but with relatively low confidence.
    \item Fig.~\ref{fig:plot-atpc-distance} shows how the accuracy drops with the increasing distance from the pedestrian to the vehicle regardless of the body size.
    \item The number of individuals from different demographic groups in the same scene affects object detection performance.
    \item The severity of weather affects different demographic groups differently.
    \item The distance between the vehicle and a pedestrian affects different demographic groups differently under different weather conditions.
\end{itemize}

In addition to demographic groups and weather conditions we considered, we speculate that factors such as the color and type of clothes, and occlusions such as wearing an umbrella or a raincoat can affect object detection performance. However, to test all these, we reiterate the importance of developing sophisticated simulators. Current simulators can have artifacts as well. For instance, we noted that simulating cloudiness results in a darker sky but a higher ambient light level in Carla, making its cloud simulation less useful for fairness analysis.

\section{Conclusions and Future Work}
We conducted an extensive analysis to identify factors that affect object detection performance. We showed the importance of considering the confidence scores of predicted bounding boxes along with other commonly used metrics. Since there are many confounding factors, we highlighted the importance of using both simulation and real-world evaluation data to thoroughly test these systems. As the next step of this research study, we aim to assess the impact of true negatives and false positives on downstream AV control tasks under varying weather conditions and demographic groups.  

\bibliographystyle{plainnat}
% \bibliographystyle{unsrtnat}
% \bibliographystyle{apalike}
% \bibliography{NLP_project_template/references}

\end{document}